\PassOptionsToPackage{table}{xcolor}

\documentclass[sigconf, screen]{acmart}

\usepackage{multirow}
\usepackage{graphicx}
\usepackage{booktabs}
\usepackage{siunitx}
\usepackage{subfigure} 
\usepackage{adjustbox}
\usepackage{makecell}
\usepackage{multirow}
\usepackage{booktabs}
\usepackage{pifont}
\usepackage{subcaption}
\usepackage{lipsum} %
\usepackage{rotating} %

\definecolor{fuchsia}{HTML}{FF0B55}

\makeatletter
\def\@authornotemark{} %
\makeatother

\usepackage[capitalise]{cleveref}  %
\renewcommand\footnotetextcopyrightpermission[1]{}

\crefname{figure}{Figure}{Figures}
\Crefname{figure}{Figure}{Figures}
\crefname{table}{Table}{Tables}
\Crefname{table}{Table}{Tables}

\usepackage{CJKutf8} %

\AtBeginDocument{%
  }

\begin{document}

\title{FormFactory: An Interactive Benchmarking Suite for Multimodal Form-Filling Agents}

\author{Bobo Li$^{\#1}$, Yuheng Wang$^{\#2}$, Hao Fei$^{*1}$, Juncheng Li$^3$, Wei Ji$^4$, Mong-Li Lee$^1$, Wynne Hsu$^1$}
\affiliation{
  \institution{$^1$National University of Singapore \quad
               $^2$Wuhan University \quad
               $^3$Zhejiang University \quad
               $^4$Nanjing University}
               \city{}
\city{\{libobo, haofei37, dcsleeml, dcshsuw\}@nus.edu.sg, yuhengwang@whu.edu.cn, junchengli@zju.edu.cn, weiji@nju.edu.cn}
\country{Project Page: 
\href{https://formfactory-ai.github.io}{\textcolor{fuchsia}{\textbf{https://formfactory-ai.github.io}}}}
}
\email{}
\renewcommand{\shortauthors}{Li et al.}

\authornote{Hao Fei is the corresponding author. $^\#$ indicates equal contribution.}

\begin{abstract}
Online form filling is a common yet labor-intensive task involving extensive keyboard and mouse interactions. 
Despite the long-standing vision of automating this process with “one click,” existing tools remain largely rule-based and lack generalizable, generative capabilities.
Recent advances in Multimodal Large Language Models (MLLMs) have enabled promising agents for GUI-related tasks in general-purpose scenarios.
However, they struggle with the unique challenges of form filling, such as flexible layouts and the difficulty of aligning textual instructions with on-screen fields.
To bridge this gap, we formally define the form-filling task and propose FormFactory—an interactive benchmarking suite comprising a web-based interface, backend evaluation module, and carefully constructed dataset. 
Our benchmark covers diverse real-world scenarios, incorporates various field formats, and simulates high-fidelity form interactions.
We conduct a comprehensive evaluation of state-of-the-art MLLMs and observe that no model surpasses 5\% accuracy, underscoring the inherent difficulty of the task.
These findings also reveal significant limitations in current models’ visual layout reasoning and field-value alignment abilities.
We hope our benchmark can serve as a stepping stone for further research into robust, practical form-filling agents. 

\end{abstract}



\maketitle

\section{Introduction}

\begin{figure*}[t]
    \centering
    \includegraphics[width=0.93\linewidth]{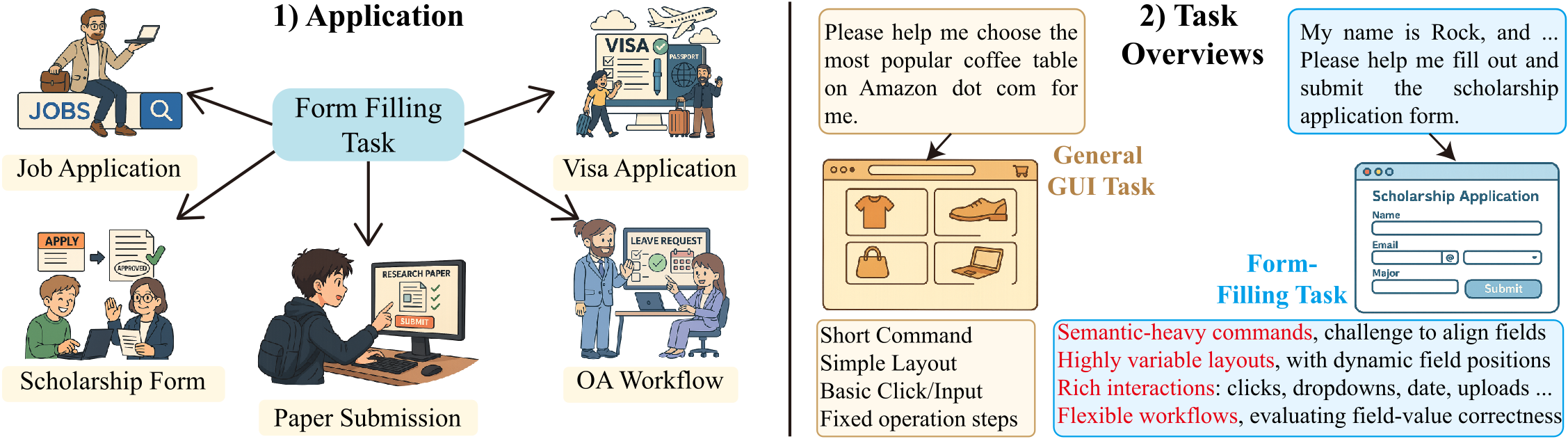}
    \caption{Overview of the form-filling task and its challenges. Compared to general GUI tasks, form-filling involves more diverse applications and demands higher semantic understanding, layout flexibility, and interaction complexity.}
    \label{fig:fig1}
\end{figure*}

Filling out online forms is a common yet tedious task in many scenarios such as job applications, academic submissions, and enterprise workflows, as illustrated in \cref{fig:fig1}.
Although users often already possess the required information—such as personal background, publication details, or leave dates—they must still carefully navigate each interface, locate the correct fields, and manually align their input.
For instance, during job-hunting season, applicants adaptively input the same resume content across a wide range of online job platforms, each with its own layout, field order, required inputs, and interaction logic.
As a result, automating the form-filling process and eliminating low-level keyboard and mouse operations has long been a highly desirable goal.

While basic solutions like Chrome’s address autofill exist, such rule-based methods are confined to static fields and often fail in dynamic or irregular form scenarios.
Some platforms offer semi-intelligent features, such as parsing uploaded resumes into structured fields on job application sites.
While helpful, these systems still struggle with accurate field-to-content alignment, particularly when dealing with nuanced distinctions or ambiguous contexts.
For example, internships may be erroneously parsed as full-time employment, distorting the applicant’s profile and demanding manual adjustments.
Accurate parsing remains unreliable even for a single form on a single site, and the diversity of real-world forms only amplifies the challenge. 
Building dedicated parsers for each platform, with its unique input types, HTML structures, and layouts, is prohibitively labor-intensive and ultimately unsustainable.
This raises a natural question: can we build a general-purpose vision agent that completes any form on any website by visually parsing the interface, retrieving relevant information, reasoning over field-value alignment, and executing the full interaction flow?

With the rise of MLLMs\cite{arlt21icml, jlbb22icml, hlvi23nips, pwqv24corr}, vision-based GUI agents have become increasingly feasible\cite{lzst24iclr, psfp23nips, whca24cvpr, kcsh24acl}.
These agents can interpret natural language instructions, perceive screen content, and execute sequences of keyboard and mouse operations to fulfill user intents.
While they have demonstrated promising performance on benchmarks for general GUI domain~\cite{tswo17icml, craa23nips, jzai24emnlp}, existing GUI agents fall short in handling form-filling tasks.
The challenges arise from three key differences:
First, form filling requires \textbf{fine-grained semantic alignment} between user-provided content (e.g., resumes) and corresponding on-screen fields. Unlike typical GUI tasks driven by explicit commands (e.g., ``install Chrome''), form filling hinges on accurately matching textual information to varied visual labels.
Second, web forms exhibit \textbf{significant structural variability}, including diverse field types, flexible layouts, and inconsistent naming conventions.
The absence of consistent patterns challenges the agent’s ability to generalize and calls for robust visual reasoning and layout comprehension. 
Table comprehension alone presents a significant challenge \cite{zhao24nips, li25www, zheng24acl} for MLLMs; form filling adds the extra complexity of aligning external content and interacting with UI elements.
Third, form filling entails \textbf{increased interaction complexity}.
Beyond basic clicks and typing, it often involves multi-step operations such as handling drop-downs, date pickers, or multi-checkbox inputs, all requiring precise action planning and interface-state awareness.
Moreover, the \textbf{non-determinism in field-click order} makes golden trajectories ambiguous, complicating supervised learning.
Collectively, these challenges place greater demands on the model’s visual grounding and semantic alignment capabilities.

Despite the practical importance and complexity of form filling, there is still no dedicated benchmark to support evaluation, let alone any well-developed models.
Existing GUI benchmarks~\cite{jzai24emnlp, craa23nips} primarily target general-purpose tasks such as app navigation or basic web operations.
However, these benchmarks fall short of addressing the specific demands of form filling, such as semantic alignment, flexible layout handling, and interaction complexity, making them unsuitable for this task.
Therefore, a reliable, interactive, and form-filling-specific benchmark is urgently needed to facilitate evaluation and further research in this domain.

To address the challenges outlined above, we propose \textbf{FormFactory}, a comprehensive benchmarking suite tailored to the form-filling task.
FormFactory consists of an interactive web platform containing 25 diverse forms drawn from realistic application scenarios, along with a paired dataset of user-provided documents or instructions and corresponding 13, 800 field-value annotations.
The suite enables MLLMs to read contextual inputs, interact with the web interface, and complete forms via click and type actions—culminating in automatic evaluation based on field-level correctness.
Our benchmark covers a wide range of domains, layouts, and field types, providing a high-fidelity simulation of real-world form-filling workflows.

We evaluate several cutting-edge MLLMs in a zero-shot setting on FormFactory. 
Despite their strong general capabilities, all models achieve under 5\% accuracy, revealing a striking mismatch between current MLLM capabilities and the complex reasoning and interaction demands of real-world form-filling tasks.

In summary, our key contributions are as follows:

\begin{itemize}
\item We pioneer the automatic form-filling task, distinguishing it from general GUI agent tasks, and establish it as a novel, practically valuable problem with real-world relevance to office automation and workload reduction.
\item We introduce a comprehensive benchmarking suite including an interactive web interface and a large-scale paired dataset, designed to support the development and evaluation of form-filling agents.
\item We conduct extensive zero-shot experiments with state-of-the-art MLLMs, revealing that form-filling remains a highly challenging task and motivating further research into vision-aligned, instruction-aware agents.
\end{itemize}

\section{Related Work}
\subsection{GUI Agent}

GUI agents~\cite{czll24corr} aim to automate human interactions with GUIs.
A range of benchmarks (including WebShop~\cite{sywt22nips}, Meta-GUI~\cite{lsmg22emnlp}, MiniWoB++~\cite{tswo17icml}, AITW~\cite{craa23nips}, Mind2Web~\cite{xdmt23nips}, WebArena~\cite{szwa24iclr}, VisualWebArena~\cite{jyve24acl}, OSWorld~\cite{txob24nips}, and AITZ~\cite{jzai24emnlp}) evaluate agents across different platforms (PC~\cite{DBLP:conf/eccv/KapoorBRKKAS24, pan2024webcanvasbenchmarkingwebagents} vs. mobile~\cite{DBLP:journals/corr/abs-2412-18426, DBLP:journals/corr/abs-2501-01149, DBLP:conf/iclr/ChenYXYCWYZLWZ025}) and interaction modes (static ~\cite{DBLP:conf/iclr/ChenH0TZZHBGCLW25, chai2025amexandroidmultiannotationexpo} vs. interactive ~\cite{bonatti2024windowsagentarenaevaluating,  DBLP:conf/nips/LinLGWYYWS24}).
These tasks typically involve generic UI navigation and can be completed using standard click operations.

On the modeling side, early approaches relied on HTML parsing and trajectory learning~\cite{lzst24iclr, pcad22icml}, but lacked generality across web structures.
Recent work has shifted toward vision-based methods using pixel inputs and large vision-language models (VLMs).
Representative systems include Pixelce-to-UI-Action~\cite{psfp23nips}, CogAgent~\cite{whca24cvpr}, Auto-GUI~\cite{zzyo24acl}, SeekClick~\cite{kcsh24acl}, and AutoWebGLM~\cite{hlaa24kdd}, which improve action grounding and decision quality via chain-of-thought prompting, visual parsing, and reinforcement learning.

However, these benchmarks and models focus on general-purpose tasks and are not designed for structured, field-intensive scenarios like form filling, which require fine-grained semantic alignment and multimodal interaction.
To address this gap, we propose FormFactory—the first benchmark dedicated to evaluating form-centric agents under realistic conditions.

\begin{table*}[!t]
\definecolor{darkgreen}{RGB}{0,128,64}
\newcommand{\cmark}{\textcolor{darkgreen}{\ding{51}}}
\centering
\fontsize{7.5}{9}\selectfont
\begin{tabular}{
 @{\hskip 2pt}c@{\hskip 2pt}
    c@{\hskip 4pt} |
    c@{\hskip 4pt}
    c@{\hskip 4pt}
    c@{\hskip 4pt}
    c@{\hskip 4pt} |
    c@{\hskip 4pt}
    c@{\hskip 4pt}
    c@{\hskip 4pt}
    c@{\hskip 4pt}
    c@{\hskip 4pt}
    c@{\hskip 4pt}
    c@{\hskip 4pt}
    c@{\hskip 4pt} |
    c@{\hskip 4pt}
}
\toprule

\makecell{Domain\\Category} & \makecell{Form Name} & \makecell{Field\\Count} & \makecell{Sample\\Count} & \makecell{Pair\\Count} & \makecell{Field\\Types} & Date & \makecell{Bin. \\Chc.} & \makecell{Drop\\down} & \makecell{Text\\Desp.} & \makecell{File\\Upload} & \makecell{Multi\\Chc.} & \makecell{Ckx.\\Input} & \makecell{Num. \\Input} & \makecell{Multi\\Page} \\

\midrule

\multirow{5}{*}{\makecell{Academic\\\& Research}} 
& Job Application for University Positions & 4 & 50 & 200 & 2 &  &  &  & \cmark &  &  &  &  &    \\
& Grant or Research Funding Application & 6 & 50 & 300 & 5 & \cmark & \cmark &  &  & \cmark &  & \cmark &  &    \\
& Paper Submission Form & 7 & 50 & 300 & 3 &  &  & \cmark &  & \cmark &  &  &  &    \\
& Student Course Registration Form & 8 & 50 & 400 & 4 &  &  & \cmark & \cmark &  & \cmark &  &  &    \\
& Scholarship Application for Students & 16 & 50 & 800 & 4 &  &  & \cmark & \cmark & \cmark &  &  &  &   \cmark \\
\midrule

\multirow{4}{*}{\makecell{Professional\\\& Business}}
& Startup Funding Application & 18 & 50 & 900 & 6 & \cmark &  & \cmark & \cmark & \cmark &  &  & \cmark &   \cmark \\
& Real Estate Rental Application & 22 & 50 & 1,100 & 6 & \cmark &  & \cmark & \cmark & \cmark &  &  & \cmark &   \cmark \\
& Educational Workshop Registration & 17 & 50 & 850 & 4 & \cmark &  & \cmark & \cmark &  &  &  &  &   \cmark \\
& Association Membership Application & 20 & 50 & 1,000 & 6 & \cmark &  & \cmark & \cmark & \cmark &  & \cmark &  &   \cmark \\
\midrule

\multirow{3}{*}{\makecell{Arts\\\& Creative}}
& Art Exhibition Submission Form & 11 & 50 & 550 & 6 &  &  & \cmark & \cmark & \cmark &  & \cmark & \cmark &   \cmark \\
& Literary Magazine Submission Form & 11 & 50 & 550 & 5 &  &  & \cmark & \cmark & \cmark &  & \cmark &  &   \cmark \\
& Conference Speaker Application Form & 14 & 50 & 700 & 6 &  & \cmark & \cmark & \cmark & \cmark &  & \cmark &  &   \cmark \\
\midrule

\multirow{2}{*}{\makecell{Technology\\\& Software}}
& Bug Reporting Form & 10 & 50 & 500 & 4 &  &  & \cmark & \cmark & \cmark &  &  &  &   \cmark \\
& IT Support Request Form & 11 & 50 & 550 & 5 &  &  & \cmark & \cmark & \cmark &  &  & \cmark &   \cmark \\
\midrule

\multirow{3}{*}{\makecell{Finance\\\& Banking}}
& Personal Loan Application Form & 7 & 50 & 350 & 3 &  &  & \cmark &  &  &  &  & \cmark &    \\
& Bank Account Opening Form & 5 & 50 & 250 & 3 & \cmark &  & \cmark &  &  &  &  &  &    \\
& Financial Planning Consultation Form & 6 & 50 & 300 & 4 & \cmark &  & \cmark & \cmark &  &  &  &  &    \\
\midrule

\multirow{3}{*}{\makecell{Healthcare\\\& Medical}}
& Patient Consent for Surgery & 8 & 50 & 400 & 3 & \cmark &  &  &  &  &  & \cmark &  &    \\
& Medical Research Study Enrollment & 8 & 50 & 400 & 4 &  &  & \cmark & \cmark &  &  &  & \cmark &    \\
& Health Insurance Claim Form & 10 & 50 & 400 & 5 & \cmark &  & \cmark & \cmark & \cmark &  &  &  &   \cmark \\
\midrule

\multirow{3}{*}{\makecell{Legal\\\& Compliance}}
& NDA Submission Form & 9 & 50 & 450 & 6 & \cmark & \cmark &  & \cmark &  &  & \cmark & \cmark &    \\
& Background Check Auth. Form & 11 & 50 & 550 & 4 & \cmark &  &  & \cmark &  &  & \cmark &  &    \\
& Contractor Onboarding Form & 14 & 50 & 700 & 6 & \cmark &  & \cmark & \cmark & \cmark &  & \cmark &  &   \cmark \\
\midrule

\multirow{2}{*}{\makecell{Construction\\\& Manufacturing}}
& Project Bid Submission Form & 13 & 50 & 650 & 5 & \cmark &  &  & \cmark & \cmark &  &  & \cmark &   \cmark \\
& Manufacturing Order Form & 13 & 50 & 650 & 5 & \cmark &  & \cmark & \cmark & \cmark &  &  &  &   \cmark \\
\midrule

Overall & / & 279 & 1,250 & 13,800 & 9 & \cmark & \cmark & \cmark & \cmark & \cmark & \cmark & \cmark & \cmark & \cmark  \\
\bottomrule
\end{tabular}

\vspace{0.5em}
\caption{
Form field statistics across domains. 
``Pair Count'' refers to the total number of field-value pairs. 
Abbreviations: Bin. Chc. = Binary Choice, Text Desp. = Text Description, Multi Chc. = Multiple Choice, Ckx. Input = Checkbox Input, Num. Input = Numeric Input. 
\cmark\ indicates the presence of a field type.
}
\vspace{-10pt}

\label{tab:form_stats}
\end{table*}

\subsection{Multimodal Large Language Model}
\label{ssec:mllm}
MLLMs extend LLMs by incorporating visual inputs, enabling joint vision-language reasoning. Notable models include CLIP~\cite{radford2021learning}, Flamingo~\cite{alayrac2022flamingo}, GPT-4V~\cite{gpt4technicalreport}, and open-source efforts such as BLIP-2~\cite{li2023blip2}, LLaVA~\cite{hlvi23nips}, and Qwen-VL~\cite{bai2023qwen}. Recent advances further explore regional understanding~\cite{chen2023shikra}, tool use~\cite{yang2023mmreact}, and modality expansion to video and 3D data~\cite{li2023videochat, hong2023llama}.

Despite their success in tasks like captioning, VQA, and instruction following, MLLMs remain limited in GUI-based scenarios that require precise spatial grounding and structured action planning. Their performance on form-centric tasks is still weak, which motivates our introduction of FormFactory, a benchmark for form-centric agents.

\begin{figure}[!tbp]
    \centering
    \includegraphics[width=0.78\linewidth]{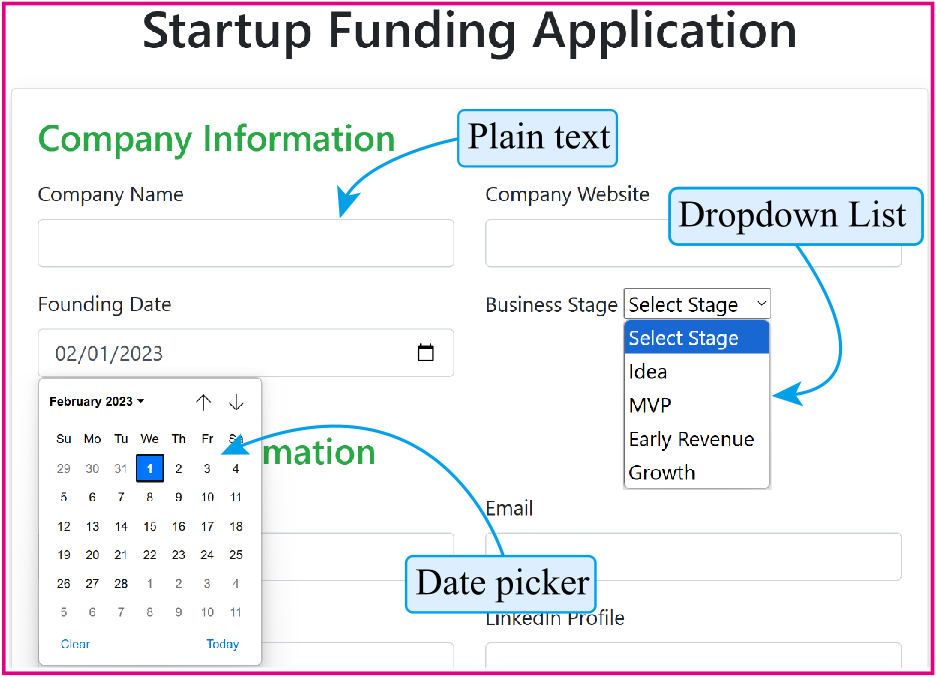}
    \caption{Example interface for a form filling task, using a startup funding application as a case study. The form contains diverse field types, including plain text inputs (e.g., company name), a date picker (e.g., founding date), and a dropdown list (e.g., business stage), illustrating the multimodal and semantically grounded nature of real-world form understanding and completion.}
    \label{fig:fig2}
\end{figure}

\begin{figure*}[t]
    \centering
    \includegraphics[width=0.98\linewidth]{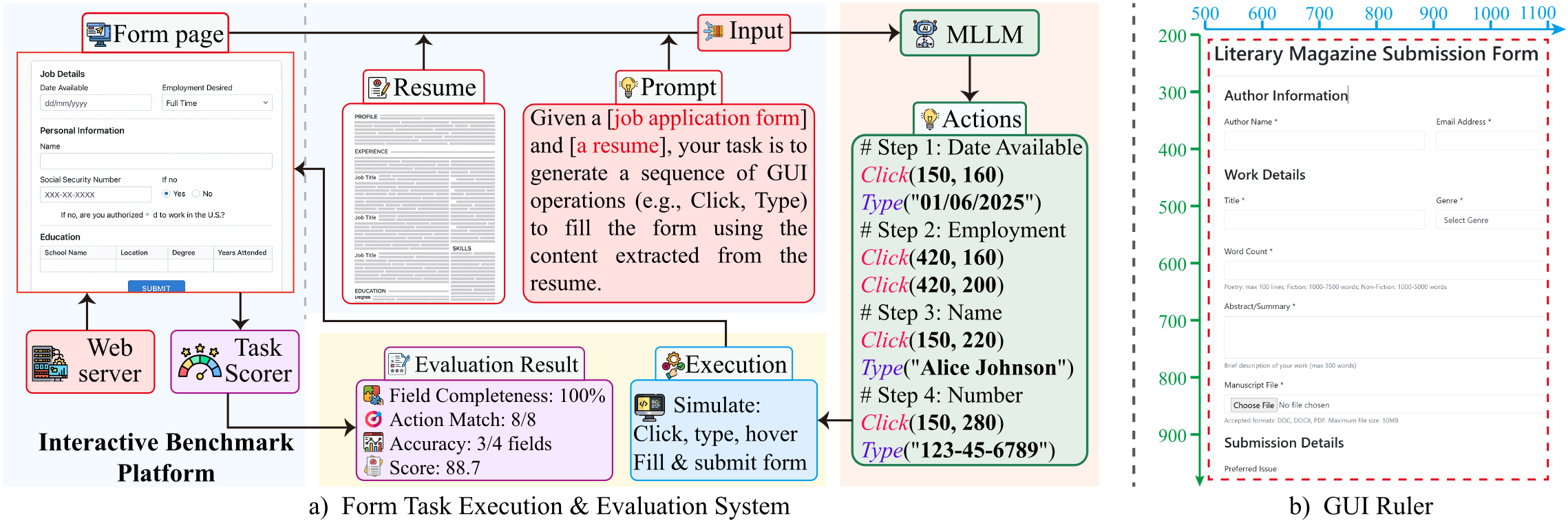}
    \caption{
    Overview of the form-filling task system. (a) The platform takes a form and a resume as input, prompting an MLLM to generate GUI actions (Click, Type) for form completion. Execution and scoring modules evaluate task performance. (b) GUI Ruler offers pixel-level reference for visual field alignment.}
    \label{fig:system}
\end{figure*}

\section{FormFactory Benchmark}

\subsection{Platform Development}
Evaluating agents on the form-filling task using real-world platforms is often impractical.
Most existing systems are tightly coupled with proprietary interfaces, lack standardized annotations, and are not designed for large-scale or automated testing, making controlled and reproducible evaluation challenging.

To overcome these limitations, we developed a dedicated simulation platform using Python and Flask.
The platform enables interactive form-filling through standard web browsers and supports automatic backend scoring based on ground-truth key-value annotations.
It is lightweight, extensible, and suitable for evaluating both human users and multimodal agents in a controlled environment.

Our benchmark covers 20 realistic forms across 8 domains, such as academia, finance, healthcare, and IT.
Each form integrates multiple field types—including text inputs, dropdowns, date pickers, checkboxes, file uploads, and numeric fields—reflecting the multimodal and compositional challenges of the task.
As illustrated in \cref{fig:fig2}, a single form may require agents to align semantics, interpret layouts, and interact across input modalities.

In addition to field diversity, we introduce stylistic variability through changes in layout, typography, and color schemes.
The platform also supports multi-page forms and modular field definitions, enabling easy expansion to new scenarios.
These design choices ensure robustness and prevent agents from overfitting to specific templates.

Overall, our platform offers a high-fidelity, flexible, and automated environment for benchmarking form-filling agents, bridging the gap between synthetic evaluation and real-world complexity.

\subsection{Dataset Collection}

We construct a dataset of input documents paired with key-value annotations, simulating the types of information typically referenced by humans when completing forms. Inputs include either document-based materials (e.g., resumes, academic papers) or free-text descriptions (e.g., for leave requests).

For each form type, we generate 50 instances using either real or LLM-synthesized data.
Most forms, such as scholarship applications and workshop registrations, are constructed in two steps.
First, we generate gold-standard field values directly from the form schema.
Then, we prompt an LLM to produce a natural language description that implicitly contains these values, simulating the type of input a human user might provide.
The generated description serves as the model’s input, while the predefined field-value pairs are used as supervision targets.
Job application forms follow a similar process, where the LLM is instructed to generate 50 detailed Markdown-style resumes containing rich personal and professional information.
For academic submission forms, we collect 50 real papers from the web and extract structured metadata—such as title, authors, and abstract—as ground truth annotations.

As shown in \cref{tab:form_stats}, this process yields a high-quality dataset of 1,250 instances spanning diverse scenarios. In total, the dataset includes 13,800 annotated field-value pairs across various field types. In addition to basic string inputs, it covers more complex types such as date pickers, dropdowns, checkboxes, multi-select options, and numerical inputs—requiring both textual understanding and layout-aware reasoning, thus posing a significant challenge to current models.

\subsection{Task Definition}

We formalize the form-filling task as a page-level sequential decision process $(c, \mathcal{E}, \mathcal{S}, \mathcal{A})$, where $c$ is the user input (e.g., a resume), $\mathcal{E}$ is the interactive web environment, $\mathcal{S}$ the visual state space (screenshots), and $\mathcal{A}$ a flat set of atomic actions.

At each step $t$, the agent observes the user input $c$ and current state $s_t \in \mathcal{S}$, then outputs a sequence of actions $\mathbf{a}_t = {a_t^1, a_t^2, \dots, a_t^k}$ to complete fields on the current page. Once done, it triggers a page-turning action (e.g., \texttt{Click(Next)}), repeating the process until submission.
The action space primarily includes:
\begin{itemize}
\item \texttt{Click(x, y)} — mouse click at pixel coordinates;
\item \texttt{Type(text)} — text input into a focused field.
\end{itemize}
Other low-level actions (e.g., \texttt{DoubleClick}, \texttt{RightClick}) may also be involved in our benchmark, but they are relatively rare.

Complex interactions (e.g., dropdowns, checkboxes, date pickers) are composed of basic actions. Our agent adopts a greedy page-wise policy, generating all applicable actions in one pass per page to reflect natural user workflows and improve efficiency.

\section{Methods}

\begin{table*}[!t]
\rowcolors{3}{gray!15}{white}
\setlength{\abovecaptionskip}{0.15cm}
\setlength{\belowcaptionskip}{0cm}
\centering
\resizebox{0.97\textwidth}{!}{%
  \begin{tabular}{l cc cc cc cc cc cc cc cc cc}
    \toprule
    \multirow{2}{*}{\textbf{Model}}
      & \multicolumn{2}{c}{\textbf{String}}
      & \multicolumn{2}{c}{\textbf{Drop-down List}}
      & \multicolumn{2}{c}{\textbf{Checkbox}}
      & \multicolumn{2}{c}{\textbf{Radio Button}}
      & \multicolumn{2}{c}{\textbf{Description}}
      & \multicolumn{2}{c}{\textbf{Date}}
      & \multicolumn{2}{c}{\textbf{Check}}
      & \multicolumn{2}{c}{\textbf{Episodic}} \\
    \cmidrule(lr){2-3} \cmidrule(lr){4-5} \cmidrule(lr){6-7} \cmidrule(lr){8-9}
    \cmidrule(lr){10-11} \cmidrule(lr){12-13} \cmidrule(lr){14-15} \cmidrule(lr){16-17}
        & \textbf{Click} & \textbf{Value}
      & \textbf{Click} & \textbf{Value}
      & \textbf{Click} & \textbf{Value}
      & \textbf{Click} & \textbf{Value}
      & \textbf{Click} & \textbf{Value}
      & \textbf{Click} & \textbf{Value}
      & \textbf{Click} & \textbf{Value}
      & \textbf{Click} & \textbf{Value} \\
    \midrule
    GPT 4o               & 2.2 & 17.5 & 0.0 & 30.7 & 0.0 & 31.3 & 0.0 & 10.0 & 8.8 & 0.8407 & 0.0 & 2.8  & 0.0 & 9.8  & 0.9 & 11.3 \\
    Gemini 2.5 Pro       & 0.9 & 98.7 & 0.0 & 99.0 & 0.0 & 76.1 & 0.0 & 52.6 & 8.1 & 0.7195 & 0.0 & 99.7 & 0.0 & 79.8 & 0.4 & 70.7 \\
    Claude 3.7 Sonnet    & 0.0 & 95.2 & 0.0 & 66.2 & 0.0 & 72.0 & 0.0 & 97.9 & 0.0 & 0.6989 & 0.0 & 99.1 & 0.0 & 55.7 & 0.0 & 58.0 \\
    Qwen-VL-Max          & 4.6 & 97.1 & 1.7 & 98.9 & 0.0 & 91.8 & 0.0 & 99.0 & 11.1 & 0.7386 & 0.0 & 98.6 & 0.0 & 71.4 & 1.1 & 72.7 \\
    Grok 3               & 0.0 & 96.2 & 0.0 & 91.2 & 0.0 & 92.4 & 0.0 & 98.1 & 5.9 & 0.7137 & 0.0 & 97.9 & 0.0 & 75.5 & 0.0 & 70.7 \\
    Doubao-vision-pro-32k & 0.0 & 94.2 & 0.0 & 89.7 & 0.0 & 38.6 & 0.0 & 96.9 & 0.0 & 0.5094 & 0.0 & 92.1 & 0.0 & 69.9 & 0.0 & 64.7 \\
    \bottomrule
  \end{tabular}%
}
\caption{
Atomic-level and episodic-level evaluation of MLLMs across different field types using Clk. and Val. metrics. GPT-4o often refuses execution despite strong capabilities, resulting in lower atomic scores. The relatively high Clk. accuracy on \textbf{Description} fields stems from their large input area, which tolerates less precise clicks. Episodic results measure end-to-end form completion accuracy.
}
\label{tab:combined}
\vspace{-10pt}
\end{table*}

\subsection{LLM-Driven Form Filling}
To enable automated execution and evaluation of form-filling tasks, we develop a lightweight MLLM-driven framework, as illustrated in \cref{fig:system} (a). The interactive benchmark system consists of three key components: a web-based form frontend, a backend task scorer, and an agent execution module.

Given a form page and an associated input document (e.g., a resume), the LLM receives a prompt instructing it to generate a sequence of GUI-level actions—such as Click(x, y) and Type(text)—to populate the form based on the extracted content. These actions are then programmatically executed on the web interface using automation tools (e.g., PyAutoGUI \cite{pyautogui}).

Once the form is submitted, the backend scorer compares the generated entries against ground-truth key-value annotations and produces a detailed evaluation report, including field-level accuracy, action match rate, and an overall task score. This framework supports scalable, fine-grained analysis of model behavior in realistic form-filling scenarios.

\subsection{Ruler-Enhanced Strategy}

While modern Multimodal LLMs exhibit impressive reasoning capabilities, they are not trained to predict pixel-level positions with precision. Directly generating absolute coordinates for GUI actions—such as clicks on form fields—can therefore be highly unreliable.
To address this limitation, we introduce a simple yet effective Ruler-Enhanced Strategy. As shown in \cref{fig:system} (b), we overlay ruler-like axes along the horizontal and vertical edges of the form screenshot. These pixel-scale markers serve as visual references that the model can use to better infer spatial relationships and generate more accurate action coordinates.

The overall task formulation and input-output format remain unchanged. We simply augment the image input with reference scales and instruct the model to utilize them when predicting Click(x, y) operations. This lightweight intervention provides auxiliary geometric context to support spatial alignment and improve the robustness of visual grounding in form-filling tasks.

\section{Experiments}
\subsection{Settings}

\subsubsection{MLLM List}
We evaluate several state-of-the-art MLLMs through their public APIs, without task-specific tuning. Each model interacts with our benchmark via an automated browser interface implemented on a Windows platform using PyAutoGUI.

Given the novelty and complexity of form-filling tasks, this evaluation setting allows us to assess the models’ inherent capabilities in visual grounding, spatial reasoning, and field-value alignment—without the confounding influence of additional training or domain adaptation.
The evaluated models include GPT-4o~\cite{openai2024gpt4ocard}, Gemini 2.5 Pro~\cite{gemini}, Claude Sonnet 3.7~\cite{claude3sonnet}, Qwen-VL-Max~\cite{qwenvlmax}, Grok 3~\cite{grok32024}, and Doubao-Vision-Pro-32k~\cite{doubao}.

\subsubsection{Evaluation Metrics}

We conduct both Atomic and Episodic evaluations. The Atomic evaluation tests model performance on individual field types—String, Drop-down List, Checkbox, Radio Button, Description, Date, and Check—to capture fine-grained interaction accuracy. In contrast, the Episodic evaluation measures end-to-end form completion, assessing the model’s ability to coordinate multiple reasoning steps in realistic workflows.

For both levels, we report two metrics: Click, evaluating correct UI element selection (using accuracy), and Value, evaluating content correctness. Value is measured by exact match accuracy (Acc), except for Description fields, which use BLEU~\cite{papineni-etal-2002-bleu} due to their generative nature.

\subsection{Main Results}

As shown in \cref{tab:combined}, most models exhibit extremely low performance on the \textbf{Click} metric—generally below 10\% across field types—indicating substantial challenges in accurate GUI interaction.
This issue is especially pronounced for structurally complex fields like checkboxes and dropdowns, where precise visual grounding and coordinate prediction are essential.
These results underscore the difficulty of our benchmark, particularly under inference-only settings without task-specific adaptation.

On the other hand, \textbf{Value} scores are relatively high.
This is partly due to our evaluation design: we consider a field correct as long as its corresponding value appears in the model’s output, regardless of whether it was actually entered into the correct UI location.
This design focuses on evaluating the model’s semantic understanding.Note that if value scores were computed based on whether the value was actually entered into the appropriate field through interaction, they would likely be significantly lower. We plan to adopt this stricter evaluation protocol in future iterations.

In summary, existing MLLMs still struggle to complete form-filling tasks reliably. Improving spatial reasoning and field alignment remains critical for enabling practical, GUI-driven agents in office automation scenarios.
\subsection{Ruler-Enhanced methods}

\begin{figure}[htbp]
    \centering
    \begin{minipage}{0.48\linewidth}
        \centering
        \includegraphics[width=\linewidth]{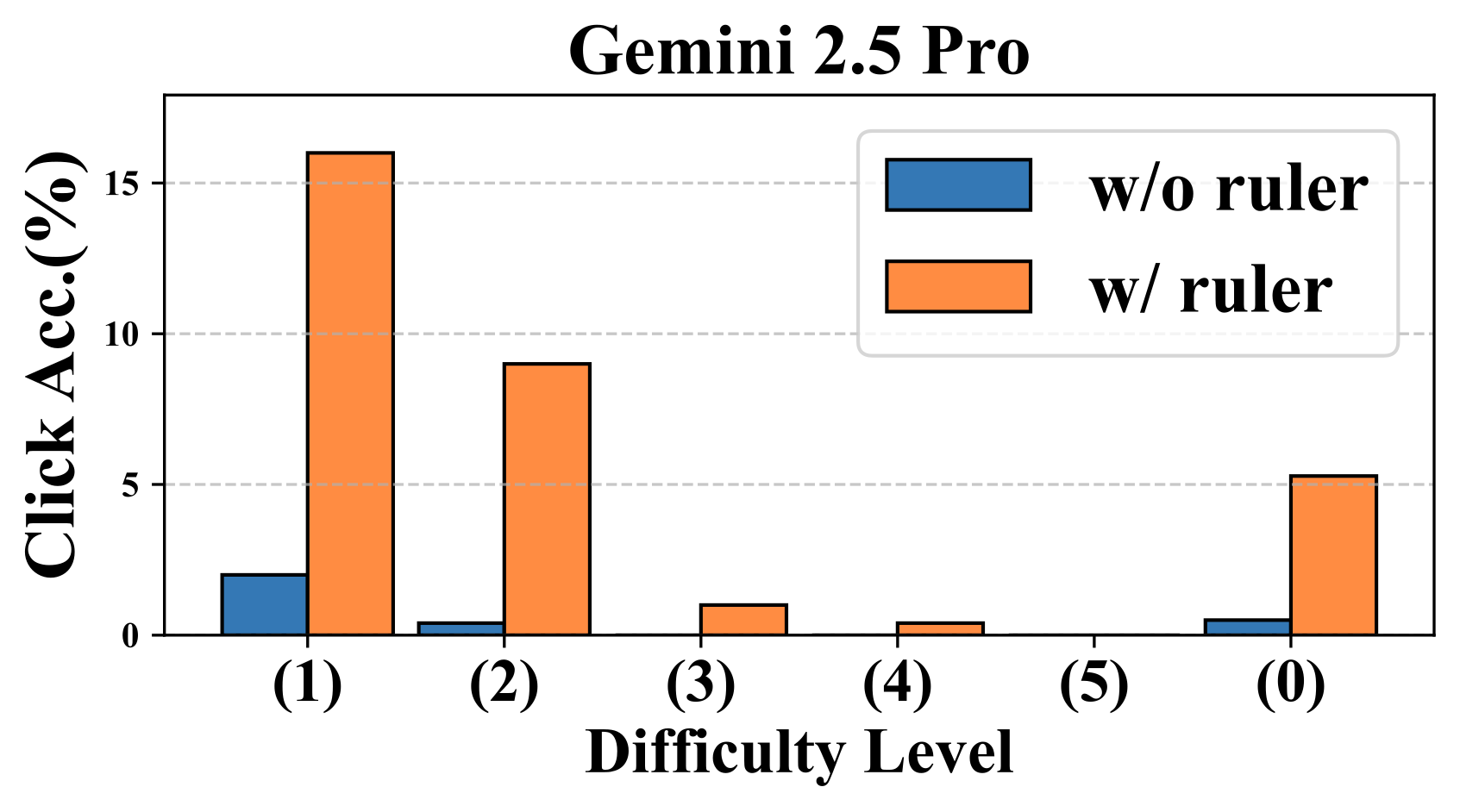}
    \end{minipage}
    \hspace{0.01\linewidth}
    \begin{minipage}{0.48\linewidth}
        \centering
        \includegraphics[width=\linewidth]{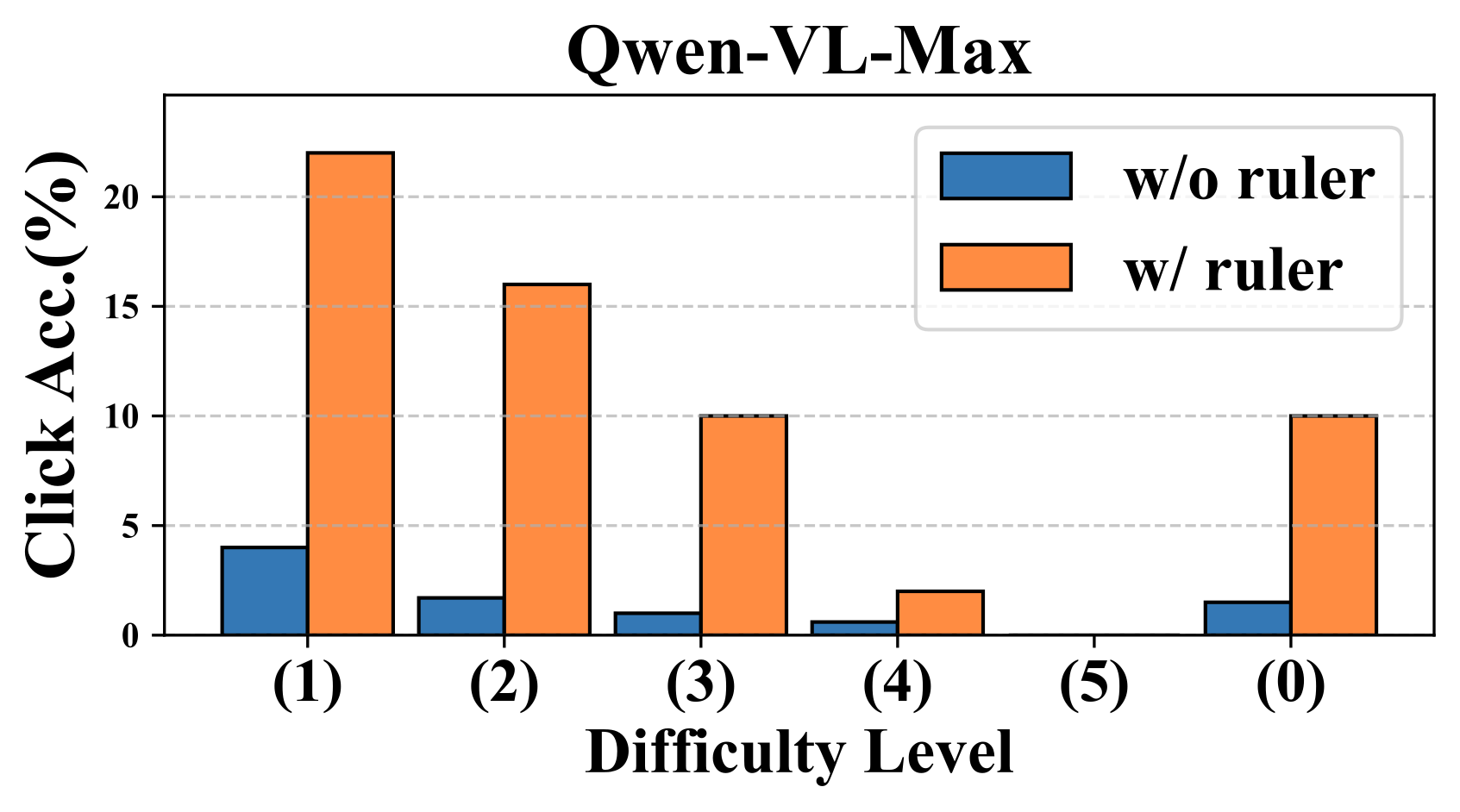}
    \end{minipage}
    \vspace{-10pt}
    \caption{
    Click accuracy with and without Ruler guidance across difficulty levels. Higher group numbers (1–5) correspond to more fields and types, indicating greater task difficulty. Group 0 reflects overall performance.
    }
    \label{fig:ruler_res}
\end{figure}

We conduct an ablation study to assess the effectiveness of the Ruler-Enhanced strategy.
As shown in \cref{fig:ruler_res}, introducing ruler guidance generally improves click accuracy, demonstrating its utility as a lightweight visual aid. The gain is more noticeable on simple forms with fewer UI elements, where spatial references help the model better align clicks with target fields. However, the improvement is marginal on complex forms requiring multiple interactions, suggesting that ruler-based guidance alone is insufficient for handling more challenging visual reasoning.

Even with ruler support, absolute click accuracy remains low, highlighting the need for more advanced strategies to improve spatial grounding and precise GUI control in form-filling tasks.

\subsection{In-depth Analysis}
\subsubsection{Value Score under Varying Field Counts}
\begin{figure}[!tbp]
    \centering
    \begin{minipage}{0.49\linewidth}
        \centering
        \includegraphics[width=\linewidth]{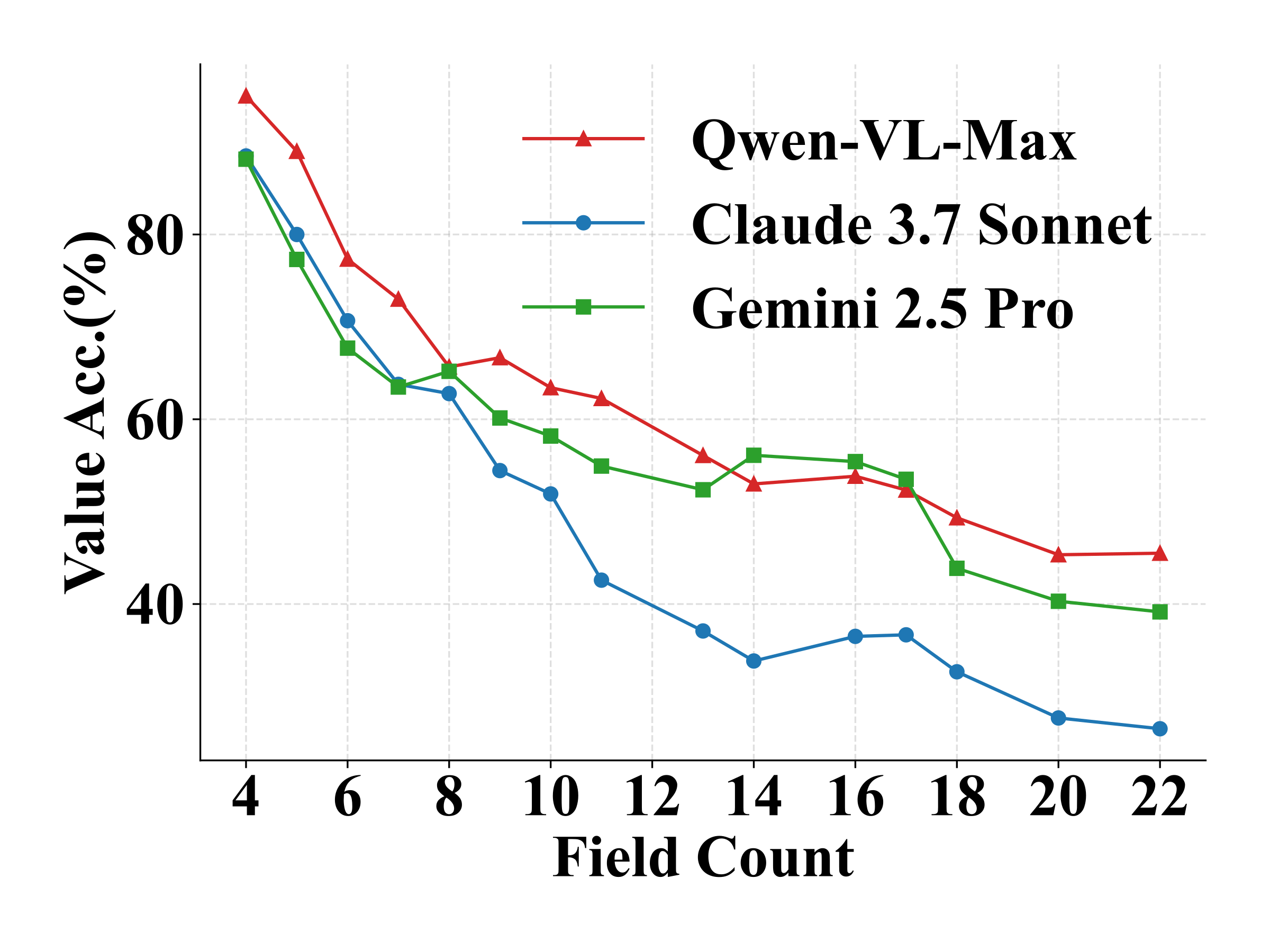}
    \end{minipage}
    \begin{minipage}{0.49\linewidth}
        \centering
        \includegraphics[width=\linewidth]{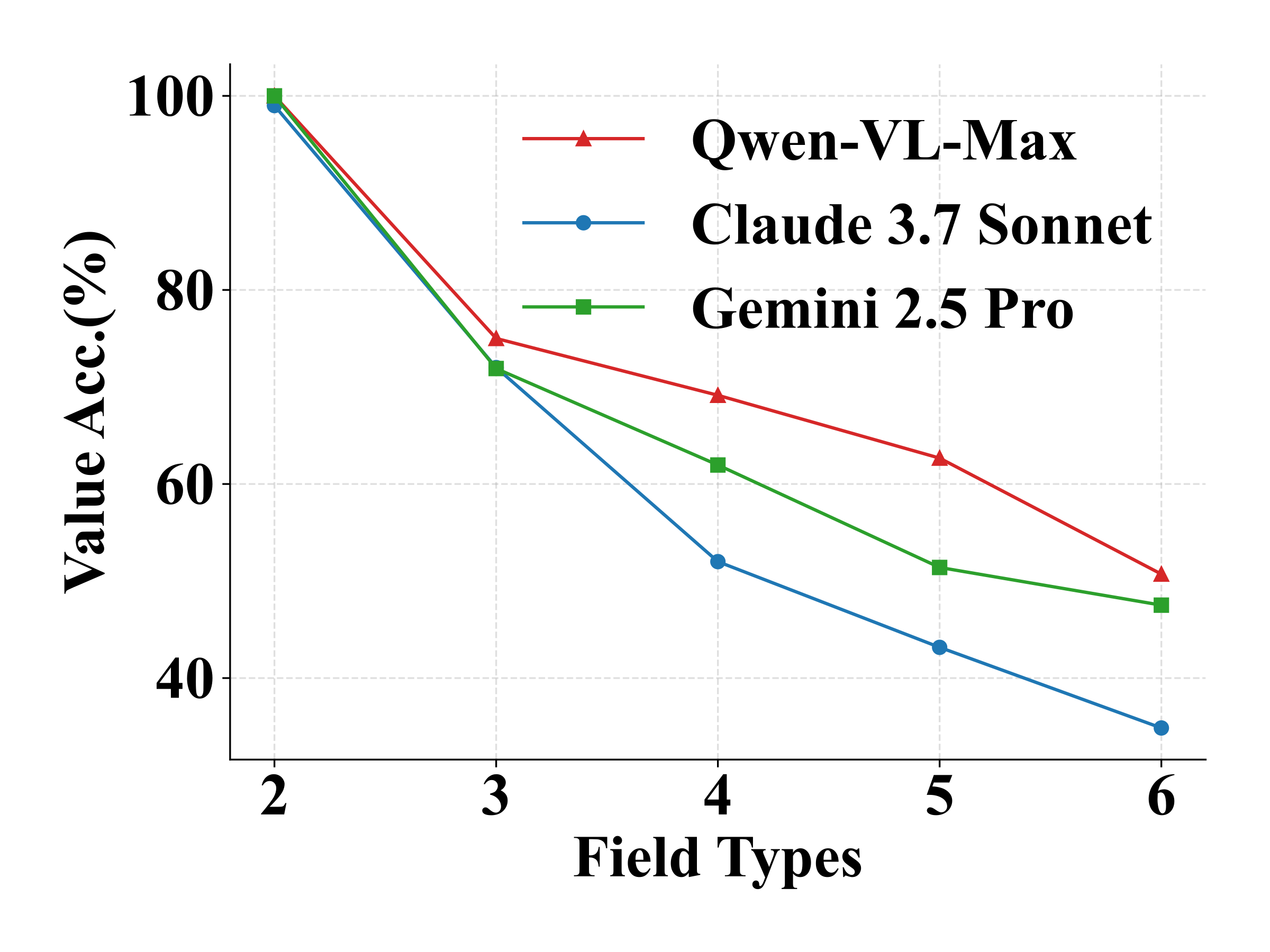}
    \end{minipage}
    \vspace{-10pt}
    \caption{Value accuracy across varying field counts (left, smoothed with a window size of 3) and field types (right).}
    \label{fig:value_field}
\end{figure}

We further analyze model performance in terms of value score under varying field complexity. As shown in \cref{fig:value_field}, all three models exhibit a clear downward trend in Value accuracy as the number of fields increases.
This reflects a natural scalability bottleneck: forms with more fields not only increase alignment difficulty but also introduce diverse field types requiring fine-grained parsing and reasoning.
Given that real-world forms often contain a large number of heterogeneous fields, this result highlights the limitations of current MLLMs and the need for more robust handling of complex form structures.

\subsubsection{Click Accuracy under Varying Field Counts}

\begin{figure}[htbp]
    \centering
    \begin{minipage}{0.49\linewidth}
        \centering
        \includegraphics[width=\linewidth]{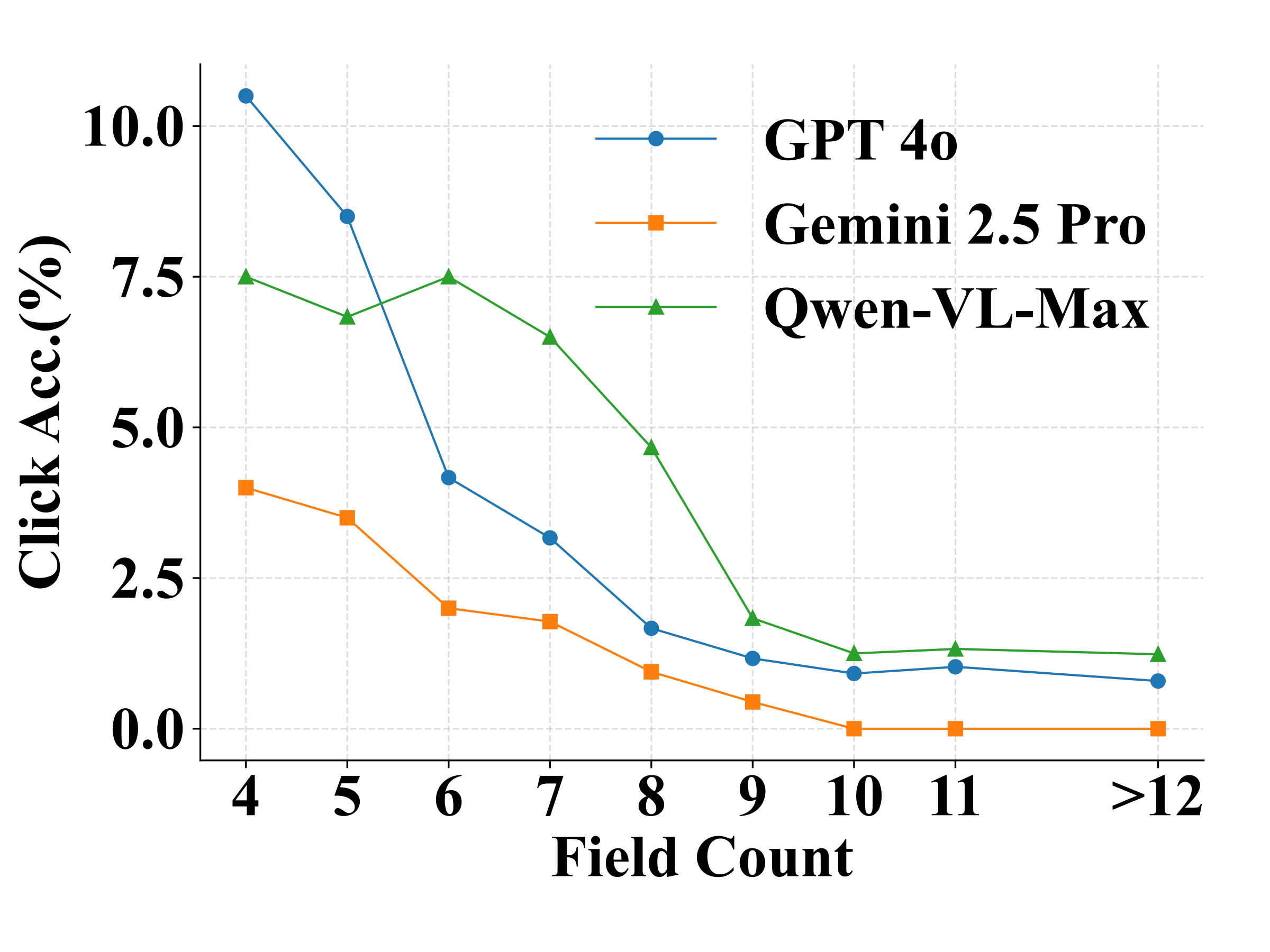}
    \end{minipage}
    \begin{minipage}{0.49\linewidth}
        \centering
        \includegraphics[width=\linewidth]{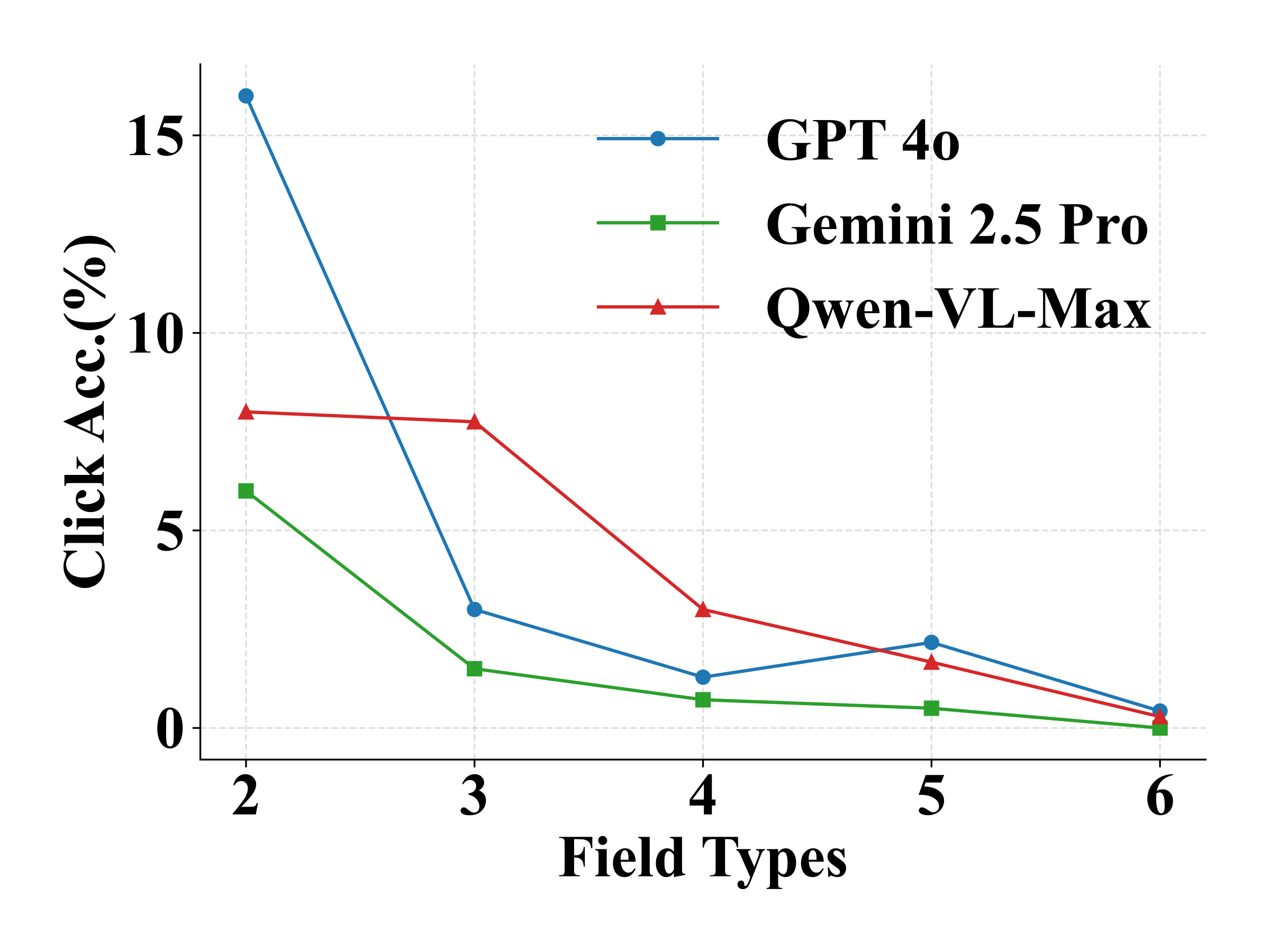}
    \end{minipage}
    \vspace{-10pt}
    \caption{Click accuracy across varying field counts (left, smoothed with a window size of 3) and field types (right).}
    \label{fig:click_trend}
\end{figure}

To complement our value-based analysis, we also examine \textbf{click accuracy} across forms of increasing field counts. As shown in \cref{fig:click_trend}, the overall trend is similar—click performance declines as field count grows, due to higher visual complexity and denser layouts.
Unlike value scoring, which applies across all field types, click accuracy is reported only for \textbf{String} and \textbf{Description} fields, as model performance on others (e.g., checkbox, dropdown) is consistently near zero. Despite this filtering, the overall click accuracy remains low across all models, underscoring the difficulty of pixel-level interaction and spatial grounding even for simple text fields.

\begin{figure}[htbp]
    \centering
    \begin{minipage}[t]{0.45\linewidth}
        \centering
        \includegraphics[width=\linewidth]{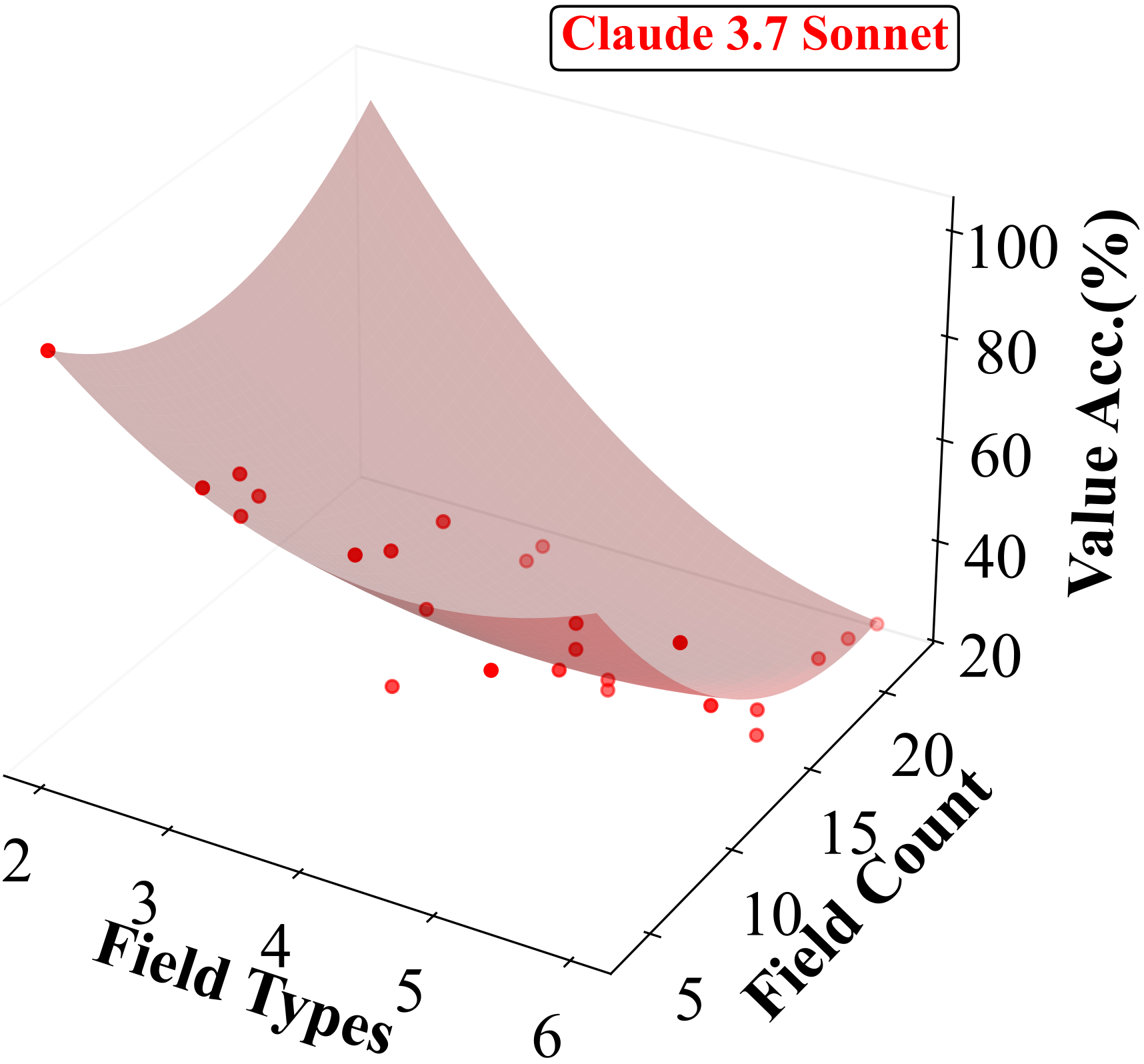}
    \end{minipage}%
    \hspace{0.01\linewidth}
    \begin{minipage}[t]{0.45\linewidth}
        \centering
        \includegraphics[width=\linewidth]{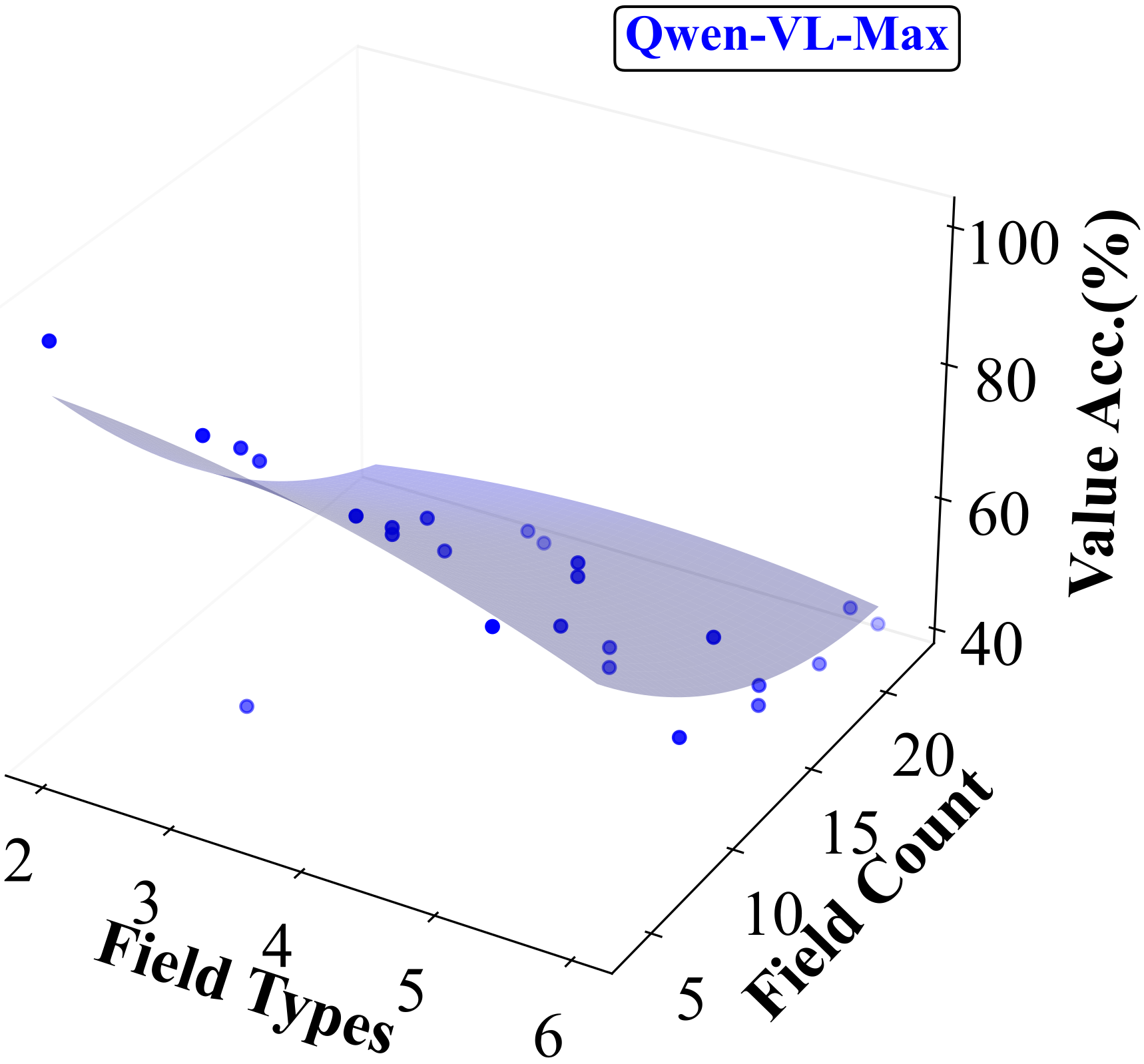}
    \end{minipage}
    \vspace{-10pt}
    \caption{
    Click accuracy under joint variation of field count and field types for two models.
    }
    \vspace{-10pt}
    \label{fig:3d}
\end{figure}

\subsubsection{Joint Effects of Field Count and Type}
While the previous sections analyzed the effects of field count and field type independently, these two factors often co-occur in real-world scenarios and jointly contribute to task complexity.
To better understand their combined impact, we conduct a 3D analysis of model performance with both variables.
As shown in \cref{fig:3d}, performance is highest when both field count and field type are low, and lowest when both are high—indicating strong compounding effects. This trend validates our benchmark design: forms with more fields and diverse field types significantly challenge the spatial reasoning, alignment, and grounding capabilities of current MLLMs.

\section{Conclusion}
In this paper, we present \textbf{FormFactory}, the first interactive benchmarking suite tailored for multimodal form-filling agents.
Our suite consists of a lightweight web platform and a diverse, realistic dataset covering a wide range of real-world scenarios. 
The benchmark is designed to simulate practical challenges, characterized by rich field types, high field counts, and complex layout variations.
We evaluate several state-of-the-art MLLMs and GUI agents under zero-shot settings.
Despite their strong performance on traditional multimodal tasks, all models exhibit severe limitations in form-filling: click success rates fall below 10\%, and end-to-end form completion rates remain under 2\%.
These results highlight that form-filling is a substantially harder task than standard GUI interaction, demanding fine-grained grounding, alignment, and reasoning.
We hope FormFactory provides a useful foundation for studying this underexplored yet practically important task.
By offering a realistic dataset, interactive interface, and standardized evaluation framework, our benchmark enables more precise diagnosis of agent limitations and facilitates systematic progress toward reliable form-filling systems.

\section{Ethical Considerations, Privacy, and Licensing}

All data in FormFactory are either synthetically generated using large language models or collected from publicly available academic papers on arXiv.
To protect author privacy, we replace all email addresses in the PDF files with randomly generated placeholders.
No personal or sensitive information is included in the released dataset.
The benchmark is released under the MIT License.

\bibliographystyle{ACM-Reference-Format}
\bibliography{main}

\end{document}